\documentclass[]{ecai}  % use option [doubleblind] for double blind submission and hiding the authors section

\usepackage{graphicx}
\usepackage{latexsym}

\usepackage{color}
\usepackage{multirow}
\usepackage[table,xcdraw]{xcolor}
\usepackage{subfigure}
\usepackage{booktabs}
\usepackage[normalem]{ulem}
\useunder{\uline}{\ul}{}
\usepackage{bbding}
\usepackage{xr}
\usepackage{amssymb}
\usepackage{amsmath}
\usepackage{float}
\usepackage{siunitx}
\ecaisubmission      % inserts page numbers. Use only for submission of paper.
\paperid{549}        % paper id for double blind submission
\begin{document}

\begin{frontmatter}

\title{Uncertainty-Encoded Multi-Modal Fusion for Robust Object Detection in Autonomous Driving}

% Single author syntax
% \author{
%     Anonymous submission
% }

\author[1]{\fnms{Yang}~\snm{Lou}}\thanks{City University of Hong Kong, email: yanglou3-c@my.cityu.edu.hk, \{qian.xu, jianwang\}@cityu.edu.hk.}
\author[2]{\fnms{Qun}~\snm{Song}}\thanks{Delft University of Technology, email: q.song-1@tudelft.nl.}
\author[1]{\fnms{Qian}~\snm{Xu}}
\author[3]{\fnms{Rui}~\snm{Tan}}\thanks{Nanyang Technological University, email: tanrui@ntu.edu.sg.}
\author[1]{\fnms{Jianping}~\snm{Wang}\vspace{-1cm}}

% \address[1]{City University of Hong Kong}
% \address[2]{Delft University of Technology}
% \address[3]{Nanyang Technological University}

\begin{abstract}
Multi-modal fusion has shown initial promising results for object detection of autonomous driving perception. However, many existing fusion schemes do not consider the quality of each fusion input and may suffer from adverse conditions on one or more sensors. While {\em predictive uncertainty} has been applied to characterize single-modal object detection performance at run time, incorporating uncertainties into the multi-modal fusion still lacks effective solutions due primarily to the uncertainty's cross-modal incomparability and distinct sensitivities to various adverse conditions.
%To improve robustness of fusion-based object detection for autonomous driving,
To fill this gap, this paper proposes {\em Uncertainty-Encoded Mixture-of-Experts} (UMoE) that explicitly incorporates single-modal uncertainties into LiDAR-camera fusion. UMoE uses individual expert network to process each sensor's detection result together with encoded uncertainty. Then, the expert networks' outputs are analyzed by a gating network to determine the fusion weights.
% \textcolor{blue}{uncertainty-encoded confidence score for fusion} 
The proposed UMoE module can be integrated into any proposal fusion pipeline. Evaluation shows that UMoE achieves a maximum of 10.67\%, 3.17\%, and 5.40\% performance gain compared with the state-of-the-art proposal-level multi-modal object detectors under extreme weather, adversarial, and blinding attack scenarios.
\end{abstract}

\end{frontmatter}

\section{Introduction}
Perception is a core subsystem of autonomous driving (AD) where onboard sensors such as LiDAR, camera, and radar are used to sense the surrounding environment. Object detection is one of the most critical perception tasks which localizes and identifies the objects of interest as important prerequisites to autonomous navigation. Recently, multi-modal fusion-based AD object detection has received enormous attention from both academia \cite{feng2020deep} and industry \cite{waymo,pony}. In particular, as LiDAR and camera provide fundamentally different and complementary information about the objects (i.e., depth and visual features), the fusion based on these two modalities has shown initial promising results for object detection \cite{huang2022multi}.

Recently, {\em predictive uncertainty} is proposed to measure the variability of model predictions under plausible inputs \cite{kendall2017uncertainties}. It has been used to measure the quality of single-modal object detection results \cite{feng2018towards,miller2019evaluating}. In the context of multi-modal fusion, we conjecture that the uncertainty regarding the sensing result in each modality is valuable to the fusion. For instance, when a sensor experiences transient interference, the resulting high uncertainty value is an important indicator for tuning down the weight of the corresponding sensing result in the fusion. However, incorporating uncertainty into fusion-based AD object detection has not received systematic study. Most existing LiDAR-camera fusion approaches \cite{sindagi2019mvx,yoo20203d,pang2020clocs} do not consider uncertainties. They may yield performance degradation when a sensor experiences sensing quality drop in certain adverse settings.
The study in \cite{feng2020leveraging} is the only work considering uncertainty in fusion-based object detection. 
However, it only considers the uncertainty of LiDAR's sensing result and does not incorporate multi-modal uncertainty into the fusion-based object detection algorithm.
Thus, it falls short of addressing the scenarios adverse on camera.

This paper aims at advancing the state of the art by designing LiDAR-camera fusion for AD object detection with each modality's predictive uncertainty incorporated. However, this turns out to be challenging, because i) there lacks informative and practical uncertainty representations, ii) the LiDAR's and camera's uncertainties, as dimensionless quantities, are not directly comparable, and iii) their sensitivities to various adverse conditions are greatly different. These properties render straightforward ways of incorporating uncertainties,  e.g., admitting raw uncertainties as fusion inputs, futile.

To address the challenges, we propose a new multi-modal fusion module called {\em Uncertainty-Encoded Mixture-of-Experts} (UMoE) for robust AD object detection.
First, UMoE applies the Monte Carlo Dropout \cite{gal2016dropout} and Direct Modeling \cite{kendall2017uncertainties} approaches to estimate each sensor's uncertainty.
Then, individual expert network is used to process each sensor's detection result with uncertainty encoded. 
Lastly, the output features of each expert network are analyzed by a gating network to determine the weights for fusion.
With the uncertainty encoding for both modalities, UMoE can retain the object detection performance or allow more graceful performance degradation when a single sensor or both sensors suffer sensing quality drops in adverse scenarios.

% Motivation Example Figure: Qualitative analysis
\begin{figure*}[ht]
\centering
\includegraphics[width=\textwidth,keepaspectratio]{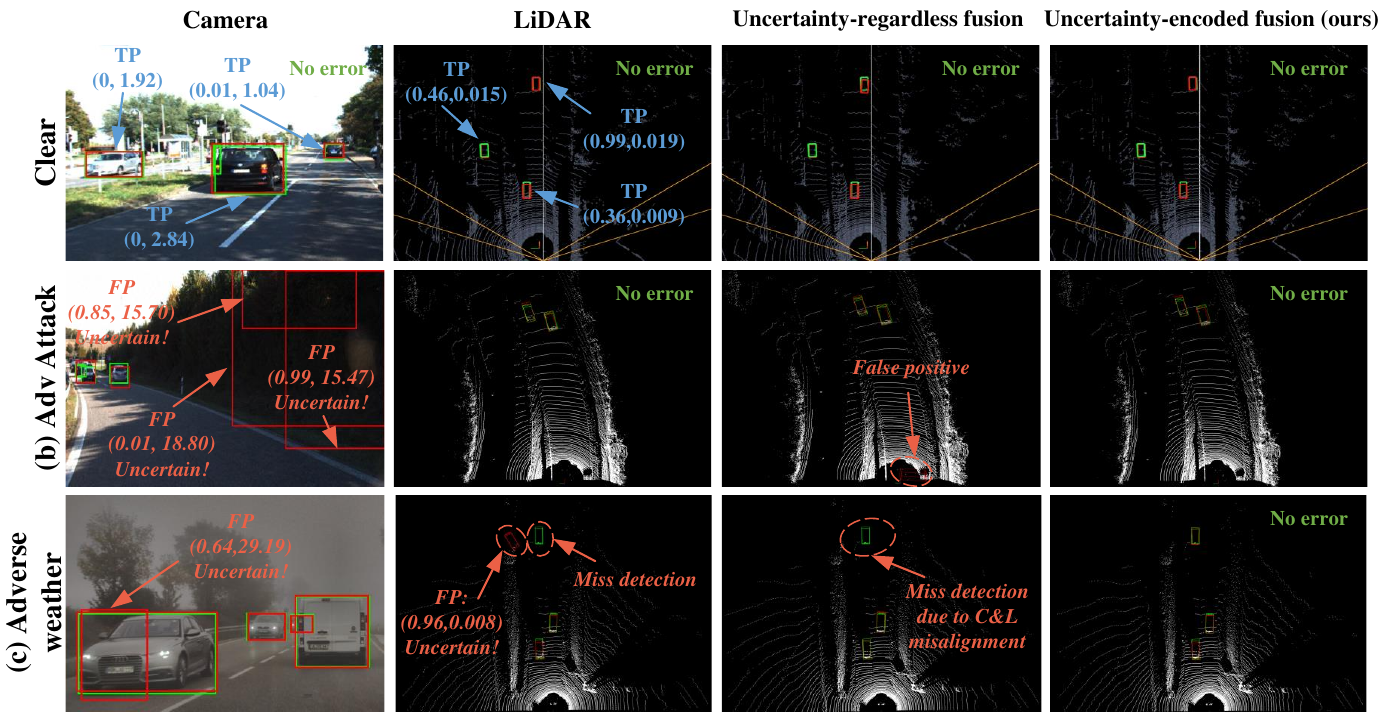}
\vspace{-1em}
\caption{Object detection by camera, LiDAR, uncertainty-regardless fusion, and our uncertainty-encoded fusion under the three driving scenes of clear, adversarial attack on camera, and adverse weather condition. Green and red bounding boxes indicate ground-truth and object detection results, respectively. TP and FP refer to true positive and false positive detection results. Number pair $(u_{cls},u_{reg})$ gives classification and regression uncertainty scores. The uncertainty-regardless fusion method is still challenged by adverse conditions, whereas our fusion method remains robust.}
% \vspace{-1em}
\label{fig:qualitative_anal}
\end{figure*}

This paper's contributions are summarized as follows:
\begin{itemize}
\setlength{\itemsep}{0pt}
\item We identify the challenges caused by the cross-modality properties of uncertainty in the multi-modal fusion design, i.e., distinct uncertainty value ranges and varied sensitivities under different adverse conditions. Based on the understanding, we encode the LiDAR and camera uncertainties into comparable scores that can be leveraged to refine detections across modalities.

\item We propose UMoE that applies encoded uncertainties to weigh and fuse the two sensing modalities for robust AD object detection. As a desirable feature, the UMoE module can be incorporated into any proposal-level fusion methods. \iffalse in a plug-and-play style \fi  To the best of our knowledge, UMoE is the first modularized mechanism incorporating multi-modal uncertainties into AD object detection.
    
\item Experiments show that UMoE outperforms advanced and state-of-the-art LiDAR-camera fusion models on real-world and synthetic datasets, including clear, snowy, foggy, adversarial, and blinding attack scenarios.
\end{itemize}

\section{Background}
{\bf Object Detection based on Camera-LiDAR Fusion: }
According to the combination stage of LiDAR and camera data representations, current methods are categorized into data-, feature-, and proposal-level fusion. 
This paper focuses on proposal-level fusion for the following reasons. First, it is difficult to quantify uncertainty in data- and feature-level fusion because existing studies estimate uncertainty based on prediction proposals.
Second, from the literature \cite{pang2020clocs,huang2022multi,pang2022fast}, proposal-level camera-LiDAR fusion achieves competitive and even superior performance compared with data- and feature-level fusion methods. 
Lastly, proposal-level fusion can easily incorporate alternative neural networks into the fusion pipeline and thus allowing easy adaptation to new object detector designs.

{\bf Predictive Uncertainty Estimation:} Given plausible inputs, predictive uncertainty measures the variability of model predictions \cite{kendall2017uncertainties}. Predictive uncertainty includes \textit{data uncertainty} due to observation noises caused by sensor measurements or environment and \textit{model uncertainty} accounting for the uncertainty of the model that can be reduced by observing enough data.
Traditionally, data and model uncertainties are modeled under the Bayesian deep learning framework. Specifically, given a data sample $\vec{x}$, the predictive uncertainty is $p(\hat{y}|\vec{x},\mathcal{D})=\int p(\hat{y}|\vec{x},\vec{w})p(\vec{w}|\mathcal{D})d\vec{w}$, where $\mathcal{D}$ denotes the training dataset, $\vec{w}$ represents the model weights, $p(\vec{w}|\mathcal{D})$ and $p(\hat{y}|\vec{x},\vec{w})$ characterize the model and data uncertainties.

Data uncertainty is often modeled by Direct Modeling \cite{gal2016uncertainty}, which assumes that the model prediction follows a probability distribution and directly predicts the parameters of such distribution using the network output layers. Model uncertainty is usually approximated using techniques such as Monte Carlo (MC) Dropout \cite{gal2016dropout} and deep ensembles \cite{lakshminarayanan2017simple}, because it is intractable to calculate the weight posterior distribution over the dataset due to vast dimensionality. 
MC Dropout interprets dropout as a Bayesian approximation of deep Gaussian process. The model uncertainty is given by performing $N$ forward passes on the same input with dropout enabled:
$p(\hat{y}|\vec{x},\mathcal{D}) \approx \frac{1}{N}\sum^N_{n=1}p(\hat{y}|\vec{x},\vec{w})$.
Deep ensemble estimates the predictive probability using an ensemble of models which have the same architecture and are trained with random initializations and data shuffled. Since deep ensemble incurs excessive memory footprint, in this paper, we adopt MC Dropout to estimate model uncertainty.

In AD, the object detector usually produces a bounding box for each detected object to describe the object location and the semantic category (e.g., car, pedestrian) with a probability score. While, the predicted bounding box regression variables are deterministic, and the probability score may not effectively characterize the classification uncertainties. Probabilistic object detectors aim to detect objects accurately and apply reliable uncertainty estimation in both classification and bounding box regression tasks, which additionally generate the {\em classification uncertainty} which  is quantified by the probability that the object belongs to the target class and the {\em regression uncertainty} which is evaluated by the variance of the probability distribution over the predicted bounding box. The latter indicates the amount of uncertainty in the position of the box corners.
Each of the classification and regression uncertainties includes data and model uncertainties.

{\bf Multi-modal Fusion based on Uncertainty:}  
Multi-modal fusion considering the inherent uncertainty of individual modalities has been explored in previous literature.
\cite{subedar2019uncertainty} estimates predictive uncertainty via variational inference across audio and visual modalities for the activity recognition task. Their approach seeks an optimal uncertainty threshold and it fuses predictive distributions that fall below this threshold using average pooling. However, it restricts its consideration to information from non-degraded modalities with low uncertainty.
Similarly, \cite{tian2020uno} merges multiple uncertainty metrics by applying a Min operation on their deviation ratios with respect to the training set. These ratios are subsequently utilized as the "temperature" for calibrating the prediction logit, and the logits are fused via the noisy-or operation. Nonetheless, this approach is primarily designed for the classification task, despite the fact that both classification and regression results serve as crucial indicators in a 3D object detection task.
\cite{chen2022multimodal} fuses detection scores from multiple modalities employing a probabilistic approach based on Bayes' rule, coupled with the weighted average of boxes based on their data uncertainties. However, the method by \cite{chen2022multimodal} necessitates conditional independence across modalities and struggles to adapt to the 3D object detection task given that the output representations from two modalities differ. 
In contrast, our method adaptively fuse LiDAR and camera detections, harnessing both classification and regression uncertainties across all levels. It effectively addresses the representation disparity between LiDAR and camera detectors and possesses the flexibility to accommodate detectors with varying cognitive abilities.

\section{Motivating Examples}
\label{sec:meas_study}
% Motivation Example: Stat Figures
\begin{figure}[h]
\centering
\includegraphics[width=0.20\textwidth]{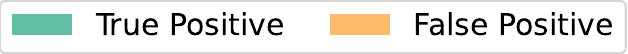}\hspace*{1.4em}
\includegraphics[width=0.20\textwidth]{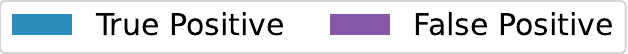}
\\
\vspace{-0.5em}
\subfigure[Camera classification]{
    \label{fig:stat_anal_camera_cls}
    \includegraphics[width=0.22\textwidth]{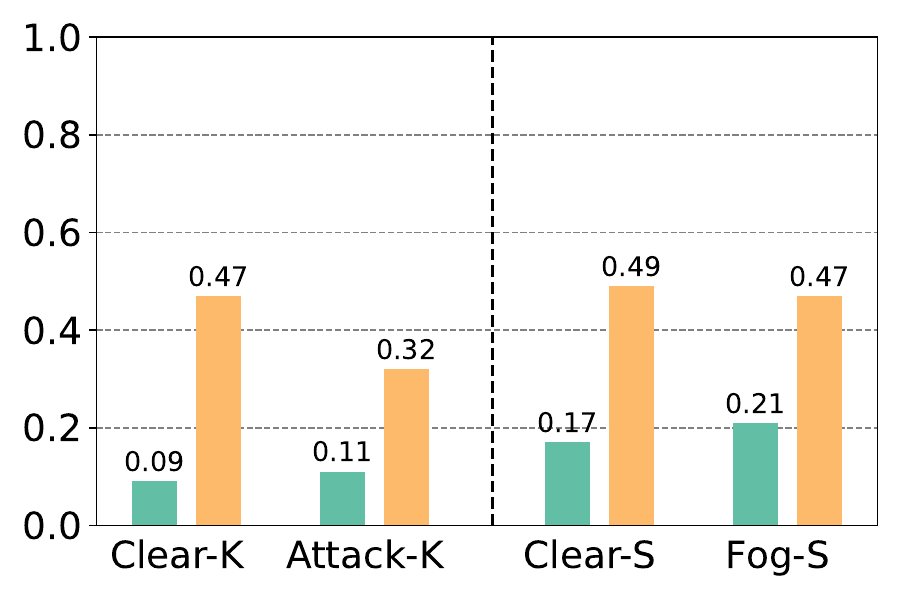}
}
\subfigure[LiDAR classification]{
    \label{fig:stat_anal_LiDAR_cls}
    \includegraphics[width=0.22\textwidth]{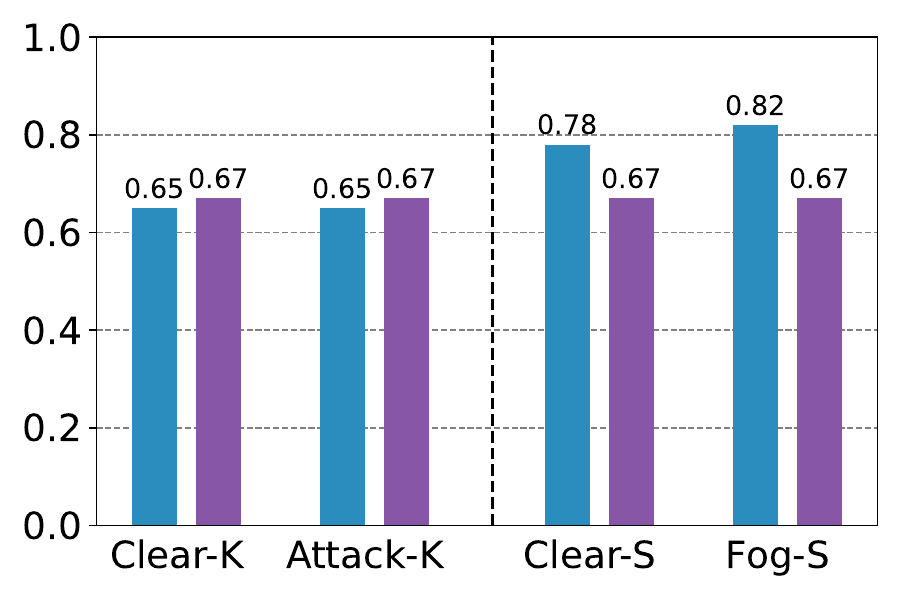}
}
\\
\subfigure[Camera regression]{
    \label{fig:stat_anal_camera_reg}
    \includegraphics[width=0.22\textwidth]{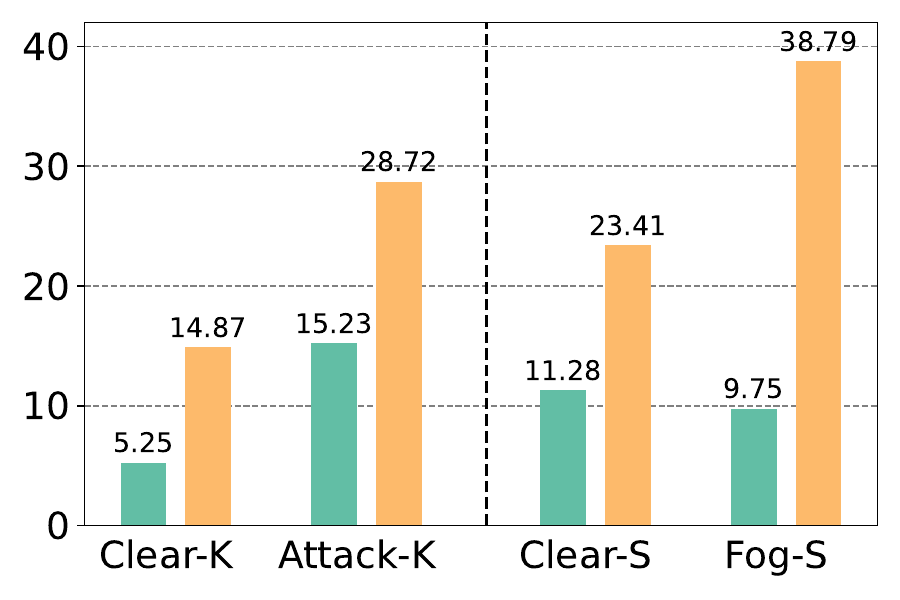}
  }
\subfigure[LiDAR regression]{
    \label{fig:stat_anal_LiDAR_reg}
    \includegraphics[width=0.22\textwidth]{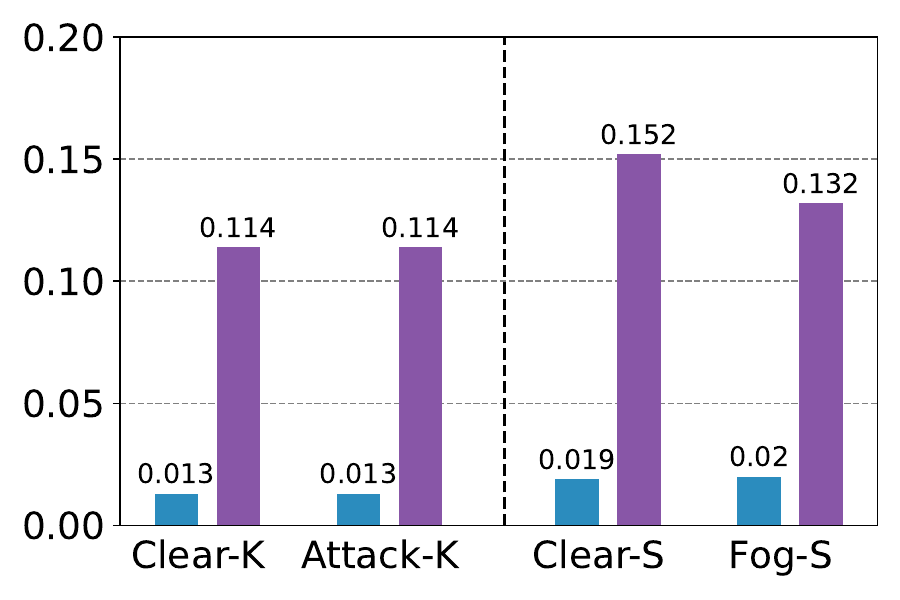}
  }
% \vspace{-0.5cm}
\caption{Average uncertainty scores in four different driving contexts: clear and synthetic adversarial attack scenes from KITTI dataset; clear and dense fog scenes from STF dataset. The vertical dashed line separates the two different datasets.}
% \vspace{-0.25cm}
\label{fig:stat_anal}
\end{figure}

For object detection in real driving scenarios, the classification and regression uncertainties are affected by variations in the environment or sensor data and the detector's cognitive ability determined by $\vec{w}$. 
For example, adverse weather and adversarial attacks can degrade the detection performance and increase the uncertainties. This section provides motivating examples to understand how the predictive uncertainties of LiDAR and camera vary in normal and adverse scenarios. We also preview the advantages of our proposed uncertainty-encoded LiDAR-camera fusion.

Fig.~\ref{fig:qualitative_anal} shows the detection results of the input frames under four representative driving contexts: (a) the clear weather scene from the KITTI dataset \cite{geiger2012we}, (b) the scene under the adversarial perturbation attack \cite{madry2018towards} against camera, (c) clear scenes, and (d) dense fog weather scenes from the STF dataset \cite{bijelic2020seeing}. We compare the detection performance of the camera-based detector RetinaNet \cite{lin2017focal}, LiDAR-based detector SECOND \cite{yan2018second}, uncertainty-regardless LiDAR-camera fusion detector CLOCs \cite{pang2020clocs}, and our uncertainty-encoded LiDAR-camera fusion detector. The detection results in the last three columns are shown in bird's-eye view.

In this section, we use Eqs.~(\ref{eq:shannon_entropy}) and (\ref{eq:total_variance}) presented later to estimate the scalar classification and regression uncertainty scores.
In a more extensive set of experiments, we compute the average classification and regression uncertainty scores of true positive and false positive detections for LiDAR and camera over datasets of the aforementioned driving scenes. False positive detections can lead to incorrect evasive actions that may cause accidents. The results are presented in Fig.~\ref{fig:stat_anal}.

Fig.~\ref{fig:qualitative_anal} and Fig.~\ref{fig:stat_anal} give four observations. First, classification and regression uncertainty scores, especially for false positives, generally increase when sensors are affected by adverse scenarios. 
Second, in the same driving scenes, false positives generate much higher classification and regression uncertainty scores than true positives. However, LiDAR tends to produce higher classification uncertainty for true positives in adverse weather condition of Fig.~\ref{fig:stat_anal_LiDAR_cls}, as a result of the lower quality and challenging nature of the dataset used. This highlights the need for utilizing both classification and regression uncertainty to address deficiencies in the dataset.
Third, camera and LiDAR have different sensitivities and cognitive abilities to the environment changes. Camera's uncertainty scores show greater volatility than LiDAR's. Lastly, camera's and LiDAR's regression uncertainty values are in different orders.

From above, predictive uncertainty is indicative of sensing performance, while camera's and LiDAR's uncertainties exhibit distinct sensitivities under different adverse scenarios. This motivates us to design a new fusion method with encoded uncertainties as part of the input. As previewed by the last column of Fig.~\ref{fig:qualitative_anal}, our method effectively exploits uncertainties for robust object detection under adverse conditions.

\section{Problem Formulation}
\label{sec:problem_formulation}
% UMoE Architecture Figure
\begin{figure*}[!t]
    \centering
    \includegraphics[width=\textwidth]{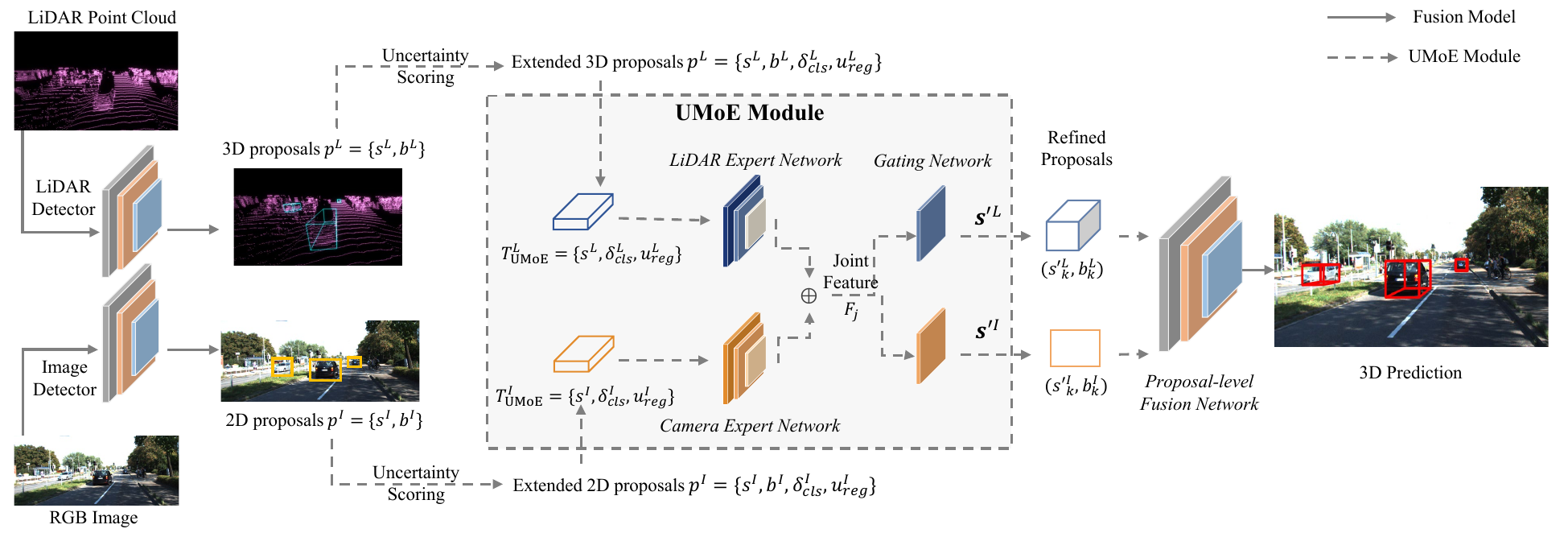}
    \vspace{-1em}
    \caption{The pipeline of the UMoE module integrated proposal-level LiDAR-camera fusion network. Dotted lines represent data flow of the UMoE module, while solid lines are for general proposal-level fusion.}
    \label{fig:UMoE_Archt}
 \end{figure*}

This paper considers fusion-based object detection using camera and LiDAR. The 3D point cloud data from LiDAR, denoted by $\vec{x}^L\in \mathbb{R}^3$, and the 2D RGB image data from camera, denoted by $\vec{x}^I \in \mathbb{R}^2$, are processed by the LiDAR detection branch $H_L$ and camera detection branch $H_I$, respectively. 
For each frame, $H_L$ produces detection $\vec{p}^L = \{\vec{p}^L_1, \ldots, \vec{p}^L_{M_L}\}$, where each $\vec{p}^L_{m_L}$ represents a proposal consisting of the 3D bounding box coordinates and the confidence score.
Similarly, $H_I$ generates detection $\vec{p}^I = \{\vec{p}^I_1, \ldots, \vec{p}^I_{M_I}\}$. 
The fusion combining $H_L$ and $H_I$ produces detection $\vec{p}=\{\vec{p}_1,\ldots,\vec{p}_K\}$, where $K={M_L} \times {M_I}$ and each element $\vec{p}_k = (\vec{p}^L_k,\vec{p}^I_k)$ is a proposal pair consisting of a LiDAR proposal and a camera proposal. 
We aim to explicitly use LiDAR's and camera's detection uncertainties in the above fusion process to derive the final detection that is robust under various adverse conditions covered by training data.

To this end, we first employ uncertainty estimation for LiDAR and camera proposals to derive LiDAR detection uncertainty $\vec{u}^{L}=\{\vec{u}_1^{L},\ldots,\vec{u}_{M_L}^{L}\}$ and camera detection uncertainty $\vec{u}^{I}=\{\vec{u}_1^{I},\ldots,\vec{u}_{M_I}^{I}\}$. Then, we aim to derive the weights that determine the importance of the two sensing modalities for each proposal pair $(\vec{p}^L_k,\vec{p}^I_k) \in \vec{p}$ based on uncertainties $\vec{u}^{L}$ and $\vec{u}^{I}$. To preserve the consistency of the input for subsequent fusion models, we implement the weights by refining the confidence score of detection result. Specifically, we aim to find the function represented by $f_u$ that maps LiDAR and camera proposals and their uncertainty to uncertainty-encoded confidence score $\vec{s'}$, i.e., $\vec{s'}= f_u(\vec{p}^{L},\vec{u}^{L},\vec{p}^{I},\vec{u}^{I})$, where $\vec{s'}=\{\vec{{s'}^L},\vec{{s'}^I}\}$. Subsequently, proposals that have their confidence scores replaced by $\vec{s'}$ can be fused to generate the final detection: 
$\vec{p}^*=f_s(\vec{p}^L,\vec{p}^I)$, where $f_s(\cdot)$ is the fusion operation or fusion network.

There are three main challenges in designing $f_u$. First, $f_u$ requires informative and practical uncertainty representations $\vec{u}^{L}$ and $\vec{u}^{I}$ as input. The representation should be distinguishable, ensuring that it assigns different values to true positives and false positives across all scenarios. Besides, in the context of AD, the uncertainty representation should be calculated without any groundtruth labels and computationally efficient.
Second, the two modalities' uncertainties respond to adverse conditions differently. For instance, in the fog scenarios of Figs.~\ref{fig:stat_anal_camera_reg} and~\ref{fig:stat_anal_LiDAR_reg}, the average regression uncertainty score for false positives from camera increases significantly, while the value for LiDAR decreases slightly.
Third, regression uncertainties computed based on 3D and 2D bounding boxes from LiDAR and camera lie in different ranges. The common 2D bounding box representation encodes top-left and bottom-right corners in camera coordinate, while 3D bounding box includes position, dimension, and rotation angle in LiDAR coordinate. The range difference of these encoded elements results in higher regression uncertainty for a 2D bounding box than that of a 3D bounding box. Our results in Fig.~\ref{fig:stat_anal_camera_reg} and~\ref{fig:stat_anal_LiDAR_reg} exhibit this issue.
The above discrepancies render straightforward ways of incorporating uncertainties, e.g., admitting raw uncertainties as fusion input, futile.

\section{Uncertainty-encoded Mixture-of-Experts (UMoE) Fusion Module}
To address the above challenges, we propose a fusion module called Uncertainty-encoded Mixture-of-Experts (UMoE) that bridges the sensor-specific detectors and the proposal-level fusion network. Fig.~\ref{fig:UMoE_Archt} illustrates the proposal-level LiDAR-camera fusion framework with our UMoE module integrated. First, LiDAR- and camera-based detectors take sensor data to generate detection proposals. With uncertainty scoring, we extend each proposal with uncertainty scores to build the UMoE input. Then, the expert network for each sensing modality extracts sensor-specific features from the UMoE input. The gating network takes the combined features generated by the preceding expert networks and generates the uncertainty-encoded confidence scores. Finally, detection proposals with updated confidence scores can be applied in the proposal-level fusion networks.

\subsection{Uncertainty Scoring}
\label{sec: uncertainty scoring}
Now we describe our uncertainty scoring approach. Denote object proposals produced by sensor-specific detectors as $\vec{p}=\{s, \vec{b}\}$, where $s$ and $\vec{b}$ denote confidence score and bounding box. By using the MC Dropout uncertainty estimation approach, each proposal $\vec{p}_k$ is assigned with uncertainty $\vec{u}_k = \{\vec{u}_{k,cls}, \vec{u}_{k,reg}\}$ consisting of the classification uncertainty $\vec{u}_{k, cls}\in \mathbb{R}^{C}$ and regression uncertainty $\vec{u}_{k,reg}\in \mathbb{R}^{B}$, where $C$ and $B$ are the numbers of classes and  elements in bounding box representation. To encode $\vec{u}_k$ into UMoE input tensor, we transform the vectors $\vec{u}_{k,cls}$ and $\vec{u}_{k,reg}$ into scalar uncertainty scores $u_{k,cls}\in\mathbb{R}$ and $u_{k,reg}\in\mathbb{R}$ as follows.

\begin{figure}[h]
\centering
\includegraphics[width=0.20\textwidth]{Figure_ECAI/legend_camera.pdf}\hspace*{1.4em}
\includegraphics[width=0.20\textwidth]{Figure_ECAI/legend_lidar.pdf}
\\
\vspace{-0.5em}
\subfigure[Camera deviation ratio]{
    \label{fig:camera_dev_ratio}
    \includegraphics[width=0.22\textwidth]{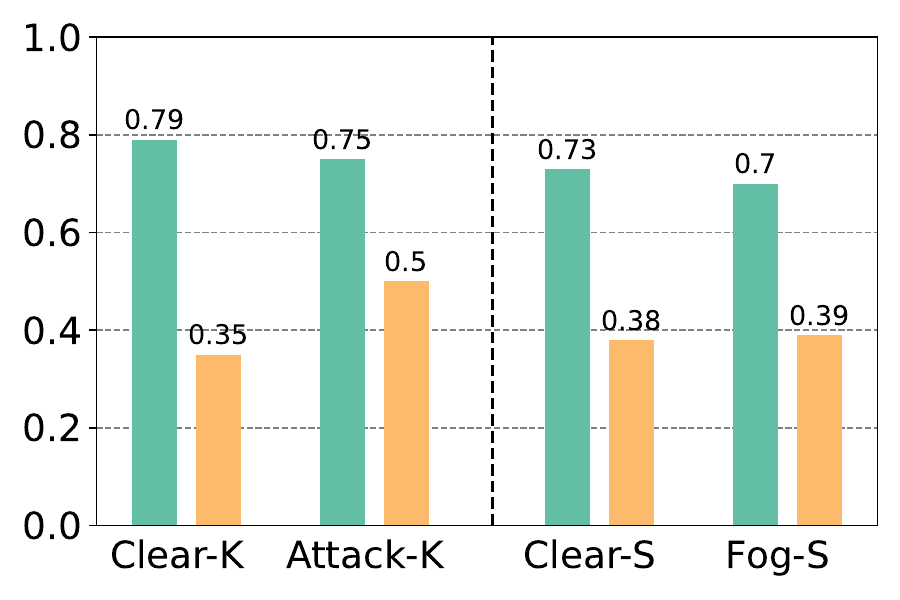}
}
\subfigure[LiDAR deviation ratio]{
    \label{fig:LiDAR_edv_ratio}
    \includegraphics[width=0.22\textwidth]{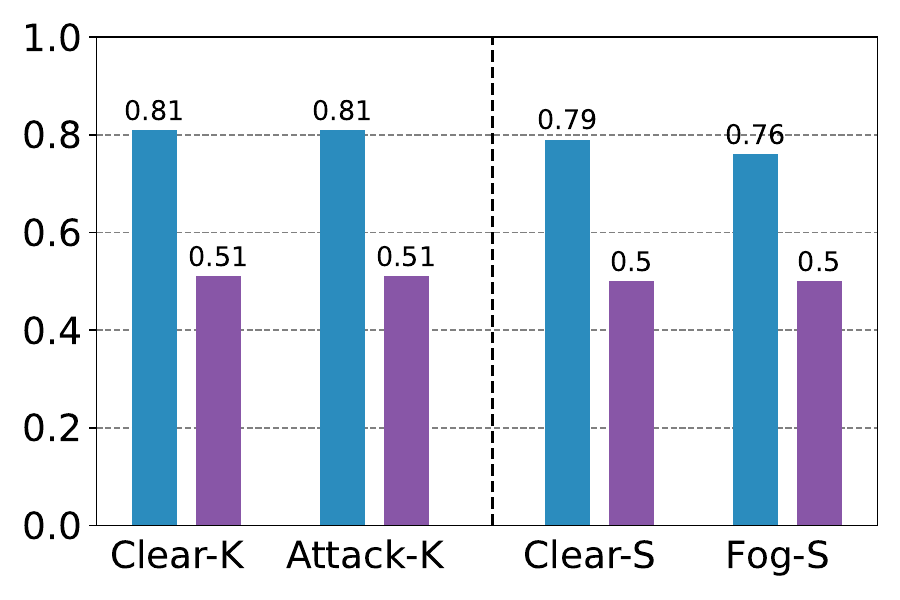}
}
% \vspace{-1.5em}
\caption{Average cls deviation ratio in the same contexts as Fig.~\ref{fig:stat_anal}.}
% \vspace{-2em}
\label{fig:dev_ratio}
\end{figure}

First, we use entropy to score classification uncertainty:
\begin{equation}\label{eq:shannon_entropy}
\textstyle
    u_{k,cls} = -\sum_{c=1}^C s_c \log s_c,
\end{equation}
where $s_c=\frac{1}{N} \sum_{n=1}^N p(\hat{y}=c|\vec{x}_k,\vec{w}_n)$ represents the average predicted classification probability of class $c$ over the $N$ forward passes of MC Dropout.
Eq.~\ref{eq:shannon_entropy} yields high classification uncertainty scores $u_{k,cls}$ for proposals with intermediate average predicted classification probability $s_c$, while demonstrating reduced values for proposals with $s_c$ at either extreme. However, the task complexity is modality-dependent, with LiDAR-based 3D detectors generally exhibit lower confidence levels compared to their camera-based 2D counterparts. Consequently, as illustrated in Fig.~\ref{fig:stat_anal_LiDAR_cls} for LiDAR classification uncertainty scores, false positives with low confidence scores have smaller average scores than true positives. 
To ensure the informativeness of the classification uncertainty score, we enhance it with a classification deviation ratio, a quantitative metric designed to evaluate the extent to which a proposal's confidence score and classification uncertainty score deviate from the true positives' distribution. Specifically:
\begin{equation}
\begin{aligned}
\delta_{k, cls} = &\frac{\mu_{u}}{\mu_{u}+\max(0, (u_{k, cls}-\mu_{u}-\sigma_{u}))} \cdot \\
&\frac{\mu_{s}}{\mu_{s}+ \max(0, -(s_{c}-\mu_{s}-\sigma_{s}))}
\end{aligned}
\end{equation}
where $\mu_{u}, \sigma_{u}, \mu_{s}, \sigma_{s}$ are the mean and standard deviation of classification uncertainty scores and the average predicted classification probability for true positives in the validation set. The computed ratio, $\delta_{k, cls}$, serves to assign larger value to proposals whose classification uncertainty score $u_{k, cls}$ and average predicted classification probability $s_{k}$ fall within the distribution of true positives, while assigning smaller values to those proposals that are out-of-distribution. As illustrated in Fig.~\ref{fig:dev_ratio}, the deviation ratio is more informative than the classification uncertainty score, as it effectively captures the differences between false positives and true positives, as well as accounting for adverse conditions.

After that, we use the total variance of $\vec{u}_{k,reg}$ to score regression uncertainty:
\begin{equation}\label{eq:total_variance}
\textstyle
    % u_{reg}^{i} = trace(\frac{1}{T}\sum_{t=1}^T \vec{b}_{\vec{w}_t^{i}}(\vec{b}_{\vec{w}_t^{i}})^{\intercal}-\vec{\mu}^{i}(\vec{\mu}^{i})^{\intercal})
    u_{k,reg} = \mathrm{tr} \left( \frac{1}{N}\sum_{n=1}^N \vec{b}_{k,n}\vec{b}_{k,n}^{\intercal}-\vec{\bar{b}}_k\vec{\bar{b}}_k^{\intercal} \right),
\end{equation}
where $\vec{\bar{b}}_k = \frac{1}{N}\sum_{n=1}^N\vec{b}_{k,n}$ is the mean bounding box coordinates over the $N$ forward passes of MC Dropout. On top of this, we adopt the Direct Modeling method for data uncertainty of bounding boxes. We assume that each regression output follows an independent Gaussian distribution and estimate the variance of these outputs. We then generate Monte Carlo samples from this distribution and add the total variance of these samples to $u_{k,reg}$. The total variance ranges in $[0,\infty]$, where a larger value indicates higher regression uncertainty. 
However, the sizes of the 2D bounding boxes in the image plane have significant influence on the values of the total variance. For instance, a closer object with a larger bounding box leads to a large $u_{k,reg}$. To achieve fair comparisons among objects at different distances, we divide $u_{k,reg}$ by the diagonal length of the averaged bounding box $\vec{\bar{b}}_k$.
We also apply standardization to $u_{k,reg}$ with mean and standard deviation of regression uncertainty calculated from the clear validation set.

With the uncertainty scores, we extend each original proposal $\vec{p}_k =\{s_k,\vec{b}_k\}$ for both LiDAR and camera to $\vec{p}_k = \{s_k,\vec{b}_k,\delta_{k, cls},u_{k,reg}\}$. For a given scene, all of the $K$ LiDAR-camera proposal pairs are used to build the tensors
$\vec{T}^{I}_\mathrm{UMoE}=\{\vec{s}^{I},\vec{\delta}_{cls}^{I},\vec{u}_{reg}^{I}\} \in \mathbb{R}^{1\times K\times 3}$, $\vec{T}^{L}_\mathrm{UMoE}=\{\vec{s}^{L},\vec{\delta}_{cls}^{L},\vec{u}_{reg}^{L}\} \in \mathbb{R}^{1\times K\times 3}$, which will be used as input to the UMoE module.

\subsection{Multi-Modal Uncertainty Fusion via UMoE}
Our UMoE is based on the Mixture of Experts (MoE) architecture \cite{jacobs1991adaptive},
which is designed to handle multiple different tasks in complex scenarios. Existing works \cite{kim2018robust,mees2016choosing} demonstrate MoE's effectiveness in multi-modal perception including LiDAR-camera fusion. The key advantage of MoE is that it contains distinct expert networks to extract features from each of the sensing modalities and then uses a gating network to combine and learn from the different extracted features to give final output. 
To find the mapping $f_{u}$, our UMoE module substitutes the traditional inputs of MoE with $\vec{T}^{I}_\mathrm{UMoE}$ and  $\vec{T}^{L}_\mathrm{UMoE}$
%and $\vec{T}^{IoU}_\mathrm{UMoE}$
to produce uncertainty-encoded confidence scores $\vec{s'}$\iffalse fusion weights\fi.
The detailed designs of our UMoE components are as follows.

% \vspace{-0.5cm}
\subsubsection{Expert networks}
As shown in the motivating example section, different sensing modalities have distinct sensitivities and value ranges for uncertainty scoring. Thus, we exploit different expert network for each sensing modality that maps input tensors to sensor-specific features for further fusion. 
Specifically, the expert network for LiDAR branch $E_{L}$ consists of a set of Residual blocks \cite{he2016deep}. The Residual Block operation is denoted by $\mathrm{ResBlock}(c_{in},c_{out},k)$, where $c_{in}, c_{out}$ are the input and output channel size and $k$ is the kernel size of 2D convolution layers inside. In $E_{L}$, we employ three Residual blocks $\mathrm{ResBlock}(3,9,(1,1))$, $\mathrm{ResBlock}(9,18,(1,1))$, and $\mathrm{ResBlock}(18,18,(1,1))$ sequentially. Similarly, the expert network for camera branch $E_{I}$ follows the same structure. 
The process can be described as:
$\vec{F}^{L} = E_{L}(\vec{T}^{L}_\mathrm{UMoE})$, $\vec{F}^{I} = E_{I}(\vec{T}^{I}_\mathrm{UMoE})$,
where $\vec{F}^{L}, \vec{F}^{I} \in R^{1\times K\times 18}$ are the feature vectors that encode confidence score and uncertainties for each proposal of the corresponding sensing modality. 

% \vspace{-0.5cm}
\subsubsection{Gating network}
The gating network concatenates features $\vec{F}^{L}$, $\vec{F}^{I}$ across all sensor modalities into a joint feature $\vec{F}_{J}$, which yields a tensor with size $1\times K\times 36$. Next, two output branches $G^{L}(\cdot)$ and $G^{I}(\cdot)$ take the same joint feature $F_{J}$ as input and predict uncertainty-encoded confidence scores $\vec{s'}^{L}$ and $\vec{s'}^{I}$, respectively. Each output branch consists of the a single Residual block $\mathrm{ResBlock}(36,1,(1,1))$. The pipeline of the gating network is defined as follows: 
$\vec{s'}^{L} = G^{L}(\vec{F}_J)$, $\vec{s'}^{I} = G^{I}(\vec{F}_J)$, where $\vec{F}_J = \vec{F}^{L} \oplus \vec{F}^{I}$.
The outputs $\vec{s'}^{L}$ and $\vec{s'}^{I}$ are used to substitute the original confidence scores $\vec{s}^L$ and $ \vec{s}^I$ of corresponding proposals based on their uncertainty scores. These refined proposals can then be input into various proposal-level fusion networks for further processing.

\begin{table*}[h]
\centering
\caption{$AP_{3D}$ on KITTI (clear), KITTIAdv (attack) and KITTIBlind (attack) datasets w/o and w/ the UMoE module.}
% \vspace{-1.5em}
\resizebox{0.85\textwidth}{!}{
\begin{tabular}{@{}ccccccccccc@{}}
\toprule
\multirow{2}{*}{Method}                                  & \multirow{2}{*}{UMoE}           & \multicolumn{3}{c}{KITTI $AP_{3D}$}                                  & \multicolumn{3}{c}{KITTIAdv $AP_{3D}$}                               & \multicolumn{3}{c}{KITTIBlind $AP_{3D}$}        \\ \cmidrule(l){3-11} 
                                                         &                                 & easy           & mod.           & hard                                & easy           & mod.           & hard                                & easy           & mod.           & hard           \\ \midrule
\multicolumn{1}{c|}{\multirow{2}{*}{CLOCs\_SecRetina}}   & \multicolumn{1}{c|}{}           & \textbf{91.61} & 81.86          & \multicolumn{1}{c|}{77.68}          & 87.77          & 77.73          & \multicolumn{1}{c|}{73.87}          & 87.89          & 78.28          & 76.24          \\
\multicolumn{1}{c|}{}                                    & \multicolumn{1}{c|}{\checkmark} & 90.25          & \textbf{81.87} & \multicolumn{1}{c|}{\textbf{79.02}} & \textbf{89.36} & \textbf{77.82} & \multicolumn{1}{c|}{\textbf{75.32}} & \textbf{90.43} & \textbf{81.80} & \textbf{79.08} \\ \midrule
\multicolumn{1}{c|}{\multirow{2}{*}{CLOCs\_PointRetina}} & \multicolumn{1}{c|}{}           & \textbf{90.42} & \textbf{81.80} & \multicolumn{1}{c|}{\textbf{78.62}} & 85.48          & 75.41          & \multicolumn{1}{c|}{73.27}          & 84.59          & 78.28          & 75.70          \\
\multicolumn{1}{c|}{}                                    & \multicolumn{1}{c|}{\checkmark} & 89.88          & 80.47          & \multicolumn{1}{c|}{77.77}          & \textbf{88.65} & \textbf{77.82} & \multicolumn{1}{c|}{\textbf{75.55}} & \textbf{89.99} & \textbf{80.31} & \textbf{77.81} \\ \bottomrule
\end{tabular}
}
\label{tab:kitti_result}
% \vspace{0.5em}
\end{table*}

\begin{table*}[]
% \vspace{-0.1cm}
\centering
\caption{$AP_{3D}$ performance on STF clear, dense fog and snow test splits w/o and w/ the UMoE module.}
% \vspace{-0.3cm}
\resizebox{0.85\textwidth}{!}{%
\begin{tabular}{@{}ccccccccccc@{}}
\toprule
\multirow{2}{*}{Method} & \multirow{2}{*}{UMoE} & \multicolumn{3}{c}{Clear $AP_{3D}$} & \multicolumn{3}{c}{Dense Fog $AP_{3D}$} & \multicolumn{3}{c}{Snow $AP_{3D}$} \\ \cmidrule(l){3-11} 
 &  & easy & mod. & hard & easy & mod. & hard & easy & mod. & hard \\ \midrule
\multicolumn{1}{c|}{\multirow{2}{*}{CLOCs\_SecRetina}} & \multicolumn{1}{c|}{} & \textbf{49.89} & 47.97 & \multicolumn{1}{c|}{\textbf{43.88}} & 31.55 & 31.92 & \multicolumn{1}{c|}{28.10} & 43.52 & 41.14 & 36.43 \\
\multicolumn{1}{c|}{} & \multicolumn{1}{c|}{\checkmark} & 49.84 & \textbf{48.41} & \multicolumn{1}{c|}{43.40} & \textbf{37.05} & \textbf{35.48} & \multicolumn{1}{c|}{\textbf{32.32}} & \textbf{47.27} & \textbf{44.25} & \textbf{39.75} \\ \midrule
\multicolumn{1}{c|}{\multirow{2}{*}{CLOCs\_PointRetina}} & \multicolumn{1}{c|}{} & \textbf{46.53} & 43.73 & \multicolumn{1}{c|}{39.73} & 28.67 & 27.80 & \multicolumn{1}{c|}{27.34} & \textbf{43.49} & 40.16 & \textbf{36.48} \\
\multicolumn{1}{c|}{} & \multicolumn{1}{c|}{\checkmark} & 46.37 & \textbf{45.46} & \multicolumn{1}{c|}{\textbf{41.63}} & \textbf{39.34} & \textbf{37.18} & \multicolumn{1}{c|}{\textbf{32.89}} & 43.04 & \textbf{40.28} & 36.20 \\ \bottomrule
\end{tabular}
}
% \vspace{0.3em}
\label{tab:stf_result}
% \vspace{0.5em}
\end{table*}

% \vspace{-0.5cm}
\subsection{Training}\label{sec:training}
We now present the training of our UMoE module. We first train the sensor-specific detectors and then fix them to train the UMoE module. For sensor-specific detectors, we add dropout layers to enable model uncertainty estimation using the MC-Dropout approach. Moreover, we follow the Direct Modeling approach to add the following loss function to the training of the sensor-specific detectors:
$\mathcal{L}_\mathrm{add}=\frac{1}{2}\exp(-\log(\vec{\sigma}^2))||\vec{b_{gt}}-\vec{b}||+\frac{1}{2}\log(\vec{\sigma}^2)$,
where $\vec{\sigma}$ is the estimated data uncertainty, $\vec{b}$ is the predicted bounding box, and $\vec{b_{gt}}$ is the corresponding ground truth. With modified sensor-specific detectors, we can train the UMoE module solely or with a proposal-level fusion network in an end-to-end manner. In this way, the UMoE module learns to produce uncertainty-encoded confidence score via supervision from final detection $\vec{p}^*$, i.e., the 3D predictions generated by the fusion network.

\section{Experiments}
This section evaluates UMoE in comparison with uncertainty-regardless LiDAR-camera fusion methods on four datasets that cover the scenarios of clear/adverse weather conditions, adversarial attack, and camera blinding attack.

\subsection{Datasets}
% \vspace{-0.5cm}
    \textbf{Clear Scenario - KITTI}:
    For sunny daytime driving scenarios, we use KITTI 3D object detection dataset \cite{geiger2012we} and follow the standard data split \cite{chen2017multi} with a 3,712 frames \emph{train} set. We randomly divide the standard val split into a \emph{val} set with 1,884 frames and a \emph{test} set with 1,885 frames. This allows us to calculate deviation ratio and enables the evaluation of subsequent attack datasets created based on \emph{test} set.
    
    \textbf{Adversarial Attack Scenario - KITTIAdv}: 
    We synthesize the adversarial attack scenario by perturbing camera images using PGD \cite{madry2018towards} method on the divided KITTI \emph{test} set 
    (See details in Appendix~\ref{sec:appendix_kittiadv}).
    % (See details in Appendix A.1\footnote{The appendix can be found online at}).
    The attack strength of PGD attack is $4/255$.
    
    \textbf{Camera Blind Attack Scenario - KITTIBlind}: We follow methods in \cite{zhang2020detecting} to generate facula with a radius of 112 pixel and overlay it on the divided KITTI \textit{test} set to form our self-synthetic KITTIBlind dataset that mimics strong light exposure affecting the camera modality (details are described in Appendix~\ref{sec:appendix_kittiblind}). 
    
    \textbf{Adverse Weather Scenario - STF}: 
    To simulate adverse weather, we adopt the STF \cite{Bijelic_2020_CVPR} dataset including clear, dense fog and snow scenarios. The STF clear weather training set, validation set and test set has 3686, 921 and 1536 scenes. Its dense fog test set has 88 scenes; snow test set has 1161 scenes. For models trained on clear weather training set, we select the snapshot with the best performance on clear validation set and evaluate it on aforementioned test splits.

\begin{table}
\centering
% \vspace{-0.5cm}
\caption{Comparison $AP_{3D}$ results of the proposed method and the state-of-the-art fusion baselines in KITTI and KITTIBlind (attack) datasets. We highlight the best performance in bold and the second best in underline.}
% \vspace{-0.2cm}
\resizebox{0.9\columnwidth}{!}{%
\begin{tabular}{@{}ccccc@{}}
\toprule
\multirow{2}{*}{Dataset} & \multirow{2}{*}{Method} & \multicolumn{3}{c}{$AP_{3D}$} \\ \cmidrule(l){3-5} 
 &  & easy & mod. & hard \\ \midrule
\multicolumn{1}{c|}{\multirow{4}{*}{KITTI}} & \multicolumn{1}{c|}{PointPainting} & 89.23 & 79.31 & 76.86 \\
\multicolumn{1}{c|}{} & \multicolumn{1}{c|}{EPNet} & \textbf{92.16} & \textbf{82.69} & \textbf{80.10} \\
\multicolumn{1}{c|}{} & \multicolumn{1}{c|}{CLOCs} & {\ul 91.61} & 81.86 & 77.68 \\ \cmidrule(l){2-5} 
\multicolumn{1}{c|}{} & \multicolumn{1}{c|}{CLOCs + UMoE} & 90.25 & {\ul 81.87} & {\ul 79.02} \\ \midrule
\multicolumn{1}{c|}{\multirow{4}{*}{KITTIBlind}} & \multicolumn{1}{c|}{PointPainting} & 87.26 & 77.20 & 74.44 \\
\multicolumn{1}{c|}{} & \multicolumn{1}{c|}{EPNet} & 87.03 & 75.38 & 73.78 \\
\multicolumn{1}{c|}{} & \multicolumn{1}{c|}{CLOCs} & {\ul 87.89} & {\ul 78.28} & {\ul 76.24} \\ \cmidrule(l){2-5} 
\multicolumn{1}{c|}{} & \multicolumn{1}{c|}{CLOCs + UMoE} & \textbf{90.43} & \textbf{81.80} & \textbf{79.08} \\ \bottomrule
\end{tabular}
}
% \vspace{-0.5em}
\label{tab:baseline_result}
% \vspace{-1cm}
\end{table}

\begin{table*}[h]
\centering
\caption{Ablation study about the effects of classification deviation ratio (DR.), regression uncertainty score (Reg.) and the MoE architecture. The best results are in bold and the second best are underlined.}
% \vspace{-0.1cm}
\resizebox{0.9\textwidth}{!}{%
\begin{tabular}{@{}ccccccccccccccc@{}}
\toprule
\multirow{2}{*}{DR.} & \multirow{2}{*}{Reg.} & \multirow{2}{*}{MoE} & \multicolumn{3}{c}{KITTIAdv} & \multicolumn{3}{c}{KITTIBlind} & \multicolumn{3}{c}{STF Snow} & \multicolumn{3}{c}{STF Dense Fog} \\ \cmidrule(l){4-15} 
 &  &  & easy & mod. & hard & easy & mod. & hard & easy & mod. & hard & easy & mod. & hard \\ \midrule
 & \checkmark & \multicolumn{1}{c|}{\checkmark} & 88.91 & 77.21 & \multicolumn{1}{c|}{74.87} & {\ul 90.07} & {\ul 81.59} & \multicolumn{1}{c|}{{\ul 79.02}} & {\ul 45.91} & {\ul 42.90} & \multicolumn{1}{c|}{{\ul 38.99}} & {\ul 36.58} & 35.00 & {\ul 32.12} \\
\checkmark &  & \multicolumn{1}{c|}{\checkmark} & {\ul 89.13} & {\ul 77.76} & \multicolumn{1}{c|}{{\ul 75.31}} & 89.96 & 81.56 & \multicolumn{1}{c|}{78.91} & 45.14 & 42.51 & \multicolumn{1}{c|}{38.56} & 34.50 & 34.65 & 30.77 \\
\checkmark & \checkmark & \multicolumn{1}{c|}{} & 85.81 & 75.57 & \multicolumn{1}{c|}{73.82} & 87.53 & 80.27 & \multicolumn{1}{c|}{78.18} & 45.59 & 42.76 & \multicolumn{1}{c|}{38.87} & 35.55 & {\ul 35.37} & 31.18 \\
\checkmark & \checkmark & \multicolumn{1}{c|}{\checkmark} & \textbf{89.36} & \textbf{77.82} & \multicolumn{1}{c|}{\textbf{75.32}} & \textbf{90.43} & \textbf{81.80} & \multicolumn{1}{c|}{\textbf{79.08}} & \textbf{47.27} & \textbf{44.25} & \multicolumn{1}{c|}{\textbf{39.75}} & \textbf{37.05} & \textbf{35.48} & \textbf{32.32} \\ \bottomrule
\end{tabular}
}
\label{tab:ablation_study}
\end{table*}

% \vspace{-0.1cm}
\subsection{Implementation}
\label{sec:implementation}
% \vspace{-0.1cm}
This section describes the implementation of our approach and the employed baselines.

\textbf{Uncertainty-regardless fusion baseline}: 
We select the CLOCs \cite{pang2020clocs}, PointPainting \cite{vora2020pointpainting} and EPNet \cite{huang2020epnet} as representative proposal-level, data-level and feature-level fusion baselines. For the CLOCs, we combine SECOND \cite{yan2018second} and RetinaNet \cite{lin2017focal}, named CLOCs\_SecRetina, and adopt PointPillar \cite{lang2019pointpillars} and RetinaNet, named CLOCs\_PointRetina as the 3D and 2D detectors. We use OpenPCDet \cite{openpcdet2020} and Detectron2 \cite{wu2019detectron2} as our 3D and 2D codebases and apply the default settings. For the CLOCs fusion network, we follow \cite{pang2022fast} that use Residual blocks instead of standard $1\times 1$ convolution layers and optimized with Adam optimizer for $20$ epochs. We employed the OneCycleLR learning rate scheduler with an initial learning rate of $6\times10^{-5}$, a maximum learning rate of $6\times10^{-4}$, and a weight decay of $0.01$. A specific description of the CLOCs fusion model structure is detailed further in Appendix~\ref{sec:appendix_imple}. For PointPainting and EPNet, we fork the original implementations without any modification.

\textbf{Uncertainty-encoded fusion}: 
We integrate the UMoE module into CLOCs\_SecRetina and CLOCs\_PointRetina as our uncertainty-encoded fusion models. To enable uncertainty estimation on sensor-specific detectors, we follow two steps to retrain 3D and 2D detectors. First, we add dropout after each DeConv2D layer for the 3D detectors and after each Conv2D layer for the detection head of the RetinaNet. The dropout rate is set at $0.1$ for all the detectors. Next, the additional loss item previously mentioned is added during training to estimate data uncertainty (refer to Appendix~\ref{sec:appendix_imple} for explicit sensor-specific detectors' training settings). During the inference of sensor-specific detectors, we perform $10$ stochastic samplings with dropout enabled, as suggested in \cite{kendall2017uncertainties}. With the retrained sensor-specific detectors fixed, we apply the same settings with the CLOCs fusion network and train our UMoE module with the fusion network in an end-to-end manner.

\vspace{-0.2cm}
\subsection{Overall Performance}
\vspace{-0.1cm}
We report our evaluation results on the most dominant class, cars, in four datasets. The Average Precision of 40 recall position with an IoU threshold of 0.7 in 3D space ($AP_{3D}$) is used as the evaluation metrics. Due to the space limitation, we present visualization figures in Appendix~\ref{sec:appendix_vis_res}.

\textbf{KITTI}: Based on evaluation results on KITTI \textit{test} set, our UMoE module maintains a satisfactory overall performance under clear scenarios. The UMoE-integrated fusion model performs comparably to uncertainty-regardless fusion in Table~\ref{tab:kitti_result}. Similar observations can be found in Table~\ref{tab:baseline_result} when comparing with state-of-the-art fusion baselines.

\textbf{KITTIAdv}: 
With numerous false positives generated from the camera-based detector due to the adversarial attack, $AP_{3D}$ of uncertainty-regardless fusion baseline drops rapidly. However, the fusion models integrated with UMoE outperform the baseline, i.e., improve $AP_{3D}$ from $87.77\%$ to $89.36\%$ on easy objects. This result may be attributed to the fact that UMoE down-weights false positive proposals with large uncertainty.

\textbf{KITTIBlind}: 
As seen in the KITTIBlind dataset results presented in Table~\ref{tab:kitti_result}, our UMoE module outperforms the uncertainty-regardless fusion baseline significantly and maintains a similar level of performance as in clear scenarios. Additionally, the $AP_{3D}$ of all state-of-the-art fusion baselines decrease with affected camera under strong light exposure in Table~\ref{tab:baseline_result}, while our module remains robust.

\textbf{STF}: We show the evaluation results for the STF dataset in Table~\ref{tab:stf_result}. LiDAR and camera degrade simultaneously in extreme weather conditions. The $AP_{3D}$ for both fusion models decreases in these scenes due to noisy LiDAR point clouds and camera images. Our proposed UMoE module significantly improves the $AP_{3D}$ under adverse weather scenarios, with a maximum increase of $10.67\%$ under dense fog weather and $3.75\%$ in snow scenes. Additionally, the UMoE-integrated fusion models achieve comparable or even better results in clear scenarios. These results demonstrate that UMoE can effectively improve robustness in extreme weather conditions.

\noindent\textbf{Statistical significant test}: 
We calculate p-values on moderate $AP_{3D}$ from 10 runs of CLOCs\_SecRetina and its baseline, which are $0.01$, \num{2.9e-14}, \num{1.4e-7}, \num{1.6e-8} in KITTIAdv, KITTIBlind, snow and dense fog scenarios. Each p-value is less than $0.05$ threshold, confirming the significance of our module's improvement.

\vspace{-0.2cm}
\subsection{Ablation Study}
To analyze the effects of uncertainty scoring and the MoE architecture, we conduct ablation studies by removing each component on the CLOCs\_SecRetina model. We report results in Table~\ref{tab:ablation_study}, including $AP_{3D}$ on KITTIAdv dataset, KITTIBlind dataset, and the snow and dense fog test sets from the STF dataset.

{\bf Encoded uncertainty:} 
To study the effectiveness of the encoded uncertainties, we remove classification deviation ratio or regression uncertainty scores from the input tensor $\vec{T}^{I}_\textrm{UMoE}$ and $\vec{T}^{L}_\textrm{UMoE}$. As shown in Table \ref{tab:ablation_study}, utilizing only the classification deviation ratio (row 2) provides relatively limited benefits, though it is particularly advantageous in adversarial attack scenes. Conversely, relying solely on the regression uncertainty score (row 1) generally results in more considerable benefits, especially in dense fog scenarios, likely attributable to its effectiveness in identifying false positives. Moreover, incorporating all components (row 4) culminates in the highest performance across all scenarios, suggesting that both uncertainties serve as crucial cues in the 3D object detection task.

{\bf MoE:}
To investigate the effectiveness of MoE, we remove this architecture and feed uncertainty scores directly to the fusion layer. The results in row 3 demonstrate that even with complete uncertainties, the model without MoE performs poorly in some adversarial attack and snow scenarios. This confirms the necessity of using MoE in handling the uncertainty differences across modalities and ranges.

\vspace{-0.2cm}
\subsection{MC-dropout runtime analysis}
This section briefly analyzes the runtime of the MC-dropout technique applied in our method. As described in Section~\ref{sec:implementation}, we perform 10 MC-dropout runs only on the detection head of sensor-specific detectors during the inference. Under these settings, the running speed is around 6 fps and 40 fps for uncertainty-encoded and uncertianty-regardless RetinaNet, respectively. The speeds are 11 fps and 24 fps for SECOND detector and 15 fps and 40 fps for PointPillar. It is worth noting that our extensive experiments show that the detection performance growth stabilizes after 5 runs. With the development of edge devices, computational redundancy can be exploited when using MC-dropout. Therefore, it will not drastically increase the cost.

\vspace{-0.2cm}
\subsection{Limitations}
Our approach encounters two primary limitations. Firstly, sensor-specific detectors may suffer slight performance degradation due to uncertainty estimation techniques such as MC-Dropout, particularly in clear scenarios, which can impact fusion performance. However, our approach can adapt advanced uncertainty estimation method to minimize such reductions. Secondly, the UMoE module can only mitigate, not eliminate, the effect of adverse scenarios. In situations of all sensors fail, our method’s enhancement
remains limited.

\vspace{-0.75em}
\section{Conclusion}
Autonomous driving is moving rapidly toward a higher level of automation in more complex environments, demanding the ability of coping with all kinds of uncertainties. This paper systematically studied how to incorporate the predictive uncertainties of individual sensors in multi-modal fusion, a fundamental task of autonomous driving perception. We score uncertainties and propose a fusion module that exploits the Mixture-of-Expert architecture to encode multi-modal uncertainties in any proposal-level fusion pipelines. Experimental results show that our module significantly improves the fusion performance in adverse scenarios. In addition to LiDAR-camera fusion, the scope of our methods can be broadened to encompass various scenarios, like incorporating additional sensors such as radar, or enhancing the LiDAR-only single-modality detection which is common in industrial-level autonomous driving systems. 
% However, the UMoE module can only mitigate, not eliminate, the effect of adverse scenarios. How to guarantee the performance enhancement when all sensors fail still remains to be explored. 
Far beyond the object detection metrics, evaluating the robustness of multi-modal fusion in various downstream tasks is interesting for future work.

\clearpage

\ack The paper is partially supported by Hong Kong Research Grant Council under GRF project 11218621.

\bibliography{ecai}

\clearpage
\setcounter{section}{0}
\renewcommand{\thesection}{\Alph{section}}
\section{Appendix}

\subsection{Datasets}
In our experiment, we evaluate on datasets with attacks and adverse weather scenarios. The adverse weather dataset STF is publicly avilable, while KITTIAdv and KITTIBlind are synthetically created by us to simulate attacks. In this section, we provide detailed descriptions of how these two datasets were constructed.

\begin{itemize}
\setlength{\itemsep}{0pt}

\item \textbf{KITTIAdv dataset}\label{sec:appendix_kittiadv} is introduced to emulate scenarios where the camera is dysfunctional by the adversarial perturbation attack. The attack is conducted by adding minute, imperceptible changes to each pixel in the image. We employ the Project Gradient Descent (PGD) method to accomplish our attack goal. Let us denote the initial perturbed image to as $I^{per}_{0}$. The attack is then implemented by updating the perturbation $\delta^{per}_{n}$ via the projected loss gradient of the camera-based detector, across multiple iteration as follows: 
\begin{equation}
    \begin{aligned}
    &\delta^{per}_n=Clip_{\epsilon}\{\alpha \times sign(\nabla_{I}) L(O_{\theta}(I^{per}_n), b^{true})\}, \\
    &I^{per}_{n+1} = I^{per}_n + \delta^{per}_n
    \end{aligned}
\end{equation}
where $\text{Clip}_{\epsilon}\{\cdot\}$ guarantees that the value falls within the $[-\epsilon, \epsilon]$, $\alpha$ is the parameter that controls the attack intensity, $\text{sign}(\cdot)$ denotes the sign function, $O_{\theta}(\cdot, \cdot)$ denotes the camera-based detector parameterized by $\theta$, $L(\cdot)$ represents the loss function of $O_{\theta}(\cdot)$, $b^{\text{true}}$ is the ground truth label, and $0 \leqslant n \leqslant N-1$. In our implementation, we have set the value range $\epsilon$ at $76$, attack intensity $\alpha$ at $1$, and the iteration number $N$ at $4$. 

\item \textbf{KITTIBlind dataset}\label{sec:appendix_kittiblind} contains fabricated scenarios in which the camera is blinded by intense light beams. To create the affected camera data, we superimpose a Gaussian facula onto the image. The radius of this facula is set to 112 pixels on a KITTI image of dimensions $1242\times 374$. For each image sample in the KITTI dataset, we randomly select a location to serve as the center point for the facula placement, within a predefined region in front of the ego vehicle. This region is specified as $[621, 745]$ for the column and $[75, 299]$ for the row. 
\end{itemize}

\subsection{Implementation}\label{sec:appendix_imple}
The implementation details of sensor-specific detectors and the proposal-level fusion network are described as follows:

For the LiDAR-based detectors, we adopt to the default settings provided in the open-source codebase, OpenPCDet, to train the SECOND and PointPillar models.
For the camera-based RetinaNet detector, we leverage another open-source codebase, Detectron2, for model training. We set an initial learning rate of $0.0025$ with a drop factor of $10$ and train the model in with a batch size of $4$ for $90,000$ iterations. During the inference phase, we set the non-maximum suppression (NMS) threshold to $0.5$ for uncertainty-regardless fusion and $0.7$ for uncertainty-encoded fusion in order to meet the requisite specifications for the proposal-level fusion network input.

For the proposal-level fusion method, we utilize the state-of-the-art model, {\em CLOCs}, to demonstrate the effectiveness of our approach. For a detailed pipeline of CLOCs: in each scene, the 3D LiDAR point clouds and 2D RGB image are first processed by their respective sensor-specific models to generate LiDAR and camera proposals. Before the NMS operation, for each pair of LiDAR and camera proposals, an input tensor $T_{CLOCs} = \{IoU, s_I , s_L, d\}$ is constructed. Here, $IoU$ denotes the Intersection over Union for 2D and 3D proposals in the image plane, $s_I$ and $s_L$ represent confidence scores, and $d$ signifies the distance of the LiDAR proposal to the ego vehicle. Subsequently, $T_{CLOCs}$ is processed by its fusion layer, which consists of four $1\times 1$ 2D convolution layers to predict refined confidence scores for more accurate 3D predictions. In \cite{pang2022fast}, authors have replaced the $1\times 1$ 2D convolution layers with residual blocks to achieve superior performance. Although they only provide a description without making the code publicly available, we have reproduced the updated CLOCs as our proposal-level fusion network in the experiment.

\subsection{Statistical Analysis}
\begin{figure}[h]
\centering
\includegraphics[width=0.20\textwidth]{Figure_ECAI/legend_camera.pdf}\hspace*{1.4em}
\includegraphics[width=0.20\textwidth]{Figure_ECAI/legend_lidar.pdf}
\\
\vspace{-0.5em}
\subfigure[Camera deviation ratio]{
    \label{fig:appendix_camera_dev}
    \includegraphics[width=0.22\textwidth]{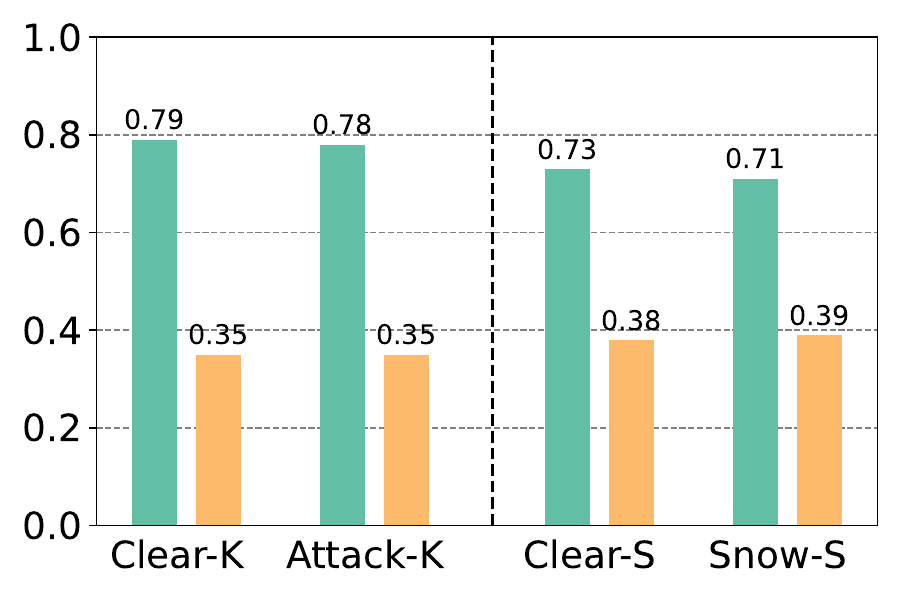}
}
\subfigure[LiDAR deviation ratio]{
    \label{fig:appendix_LiDAR_dev}
    \includegraphics[width=0.22\textwidth]{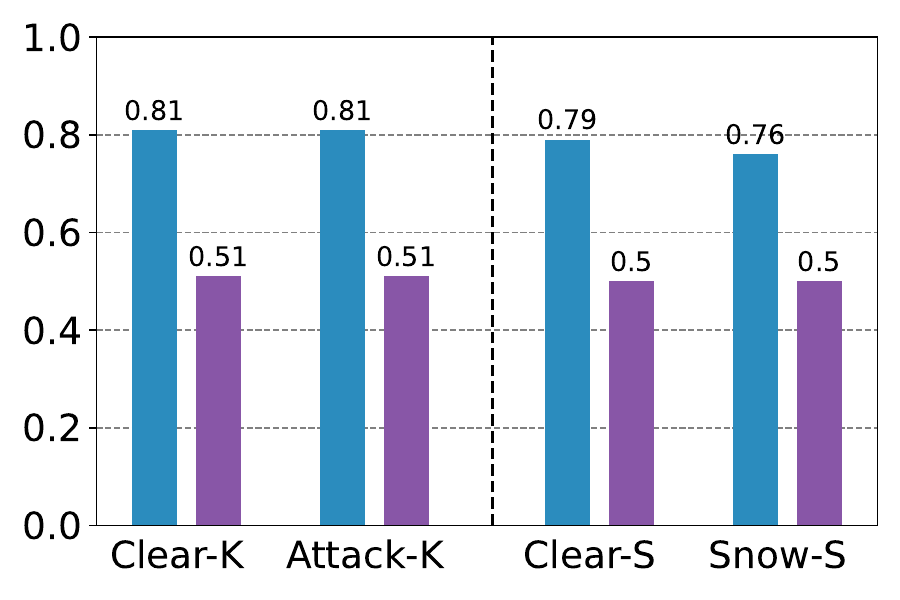}
}
\\
\subfigure[Camera regression]{
    \label{fig:appendix_camera_reg}
    \includegraphics[width=0.22\textwidth]{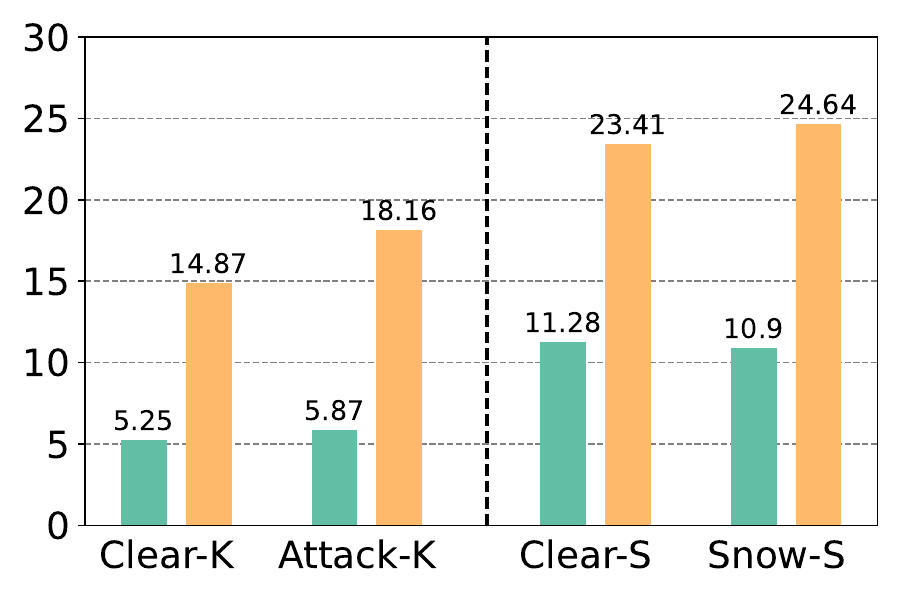}
  }
\subfigure[LiDAR regression]{
    \label{fig:appendix_LiDAR_reg}
    \includegraphics[width=0.22\textwidth]{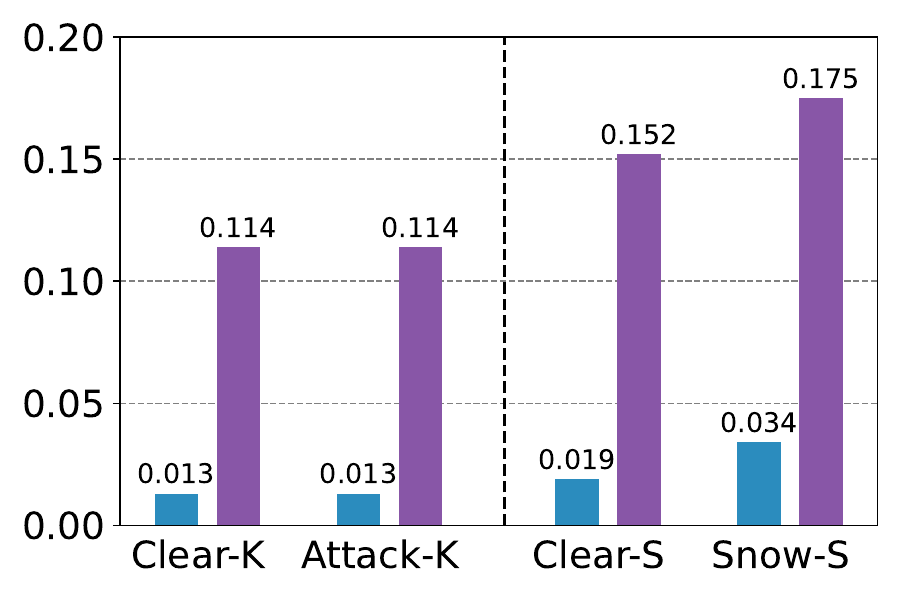}
  }
\vspace{-0.5em}
\caption{Average classification deviation ratios and regression uncertainty scores in blind attack and snow driving contexts complementing Fig.~\ref{fig:stat_anal}. Clear scenes are included for comparison.}
\label{fig:stat_anal_appendix}
\vspace{-1em}
\end{figure}

Fig.~\ref{fig:stat_anal_appendix} provides statistical analysis of classification deviation ratio and regression uncertainty score for the KITTIBlind dataset and snow scenes from STF dataset. Similar to the observations in Figures \ref{fig:stat_anal} and \ref{fig:dev_ratio}, these scores serve as discriminative indicators between true positives and false positives, and exhibit varying sensitivities to environment changes.

% Stat

\subsection{Visualization}\label{sec:appendix_vis_res}

\begin{figure*}[t]
\centering
\includegraphics[width=0.9\textwidth,keepaspectratio]{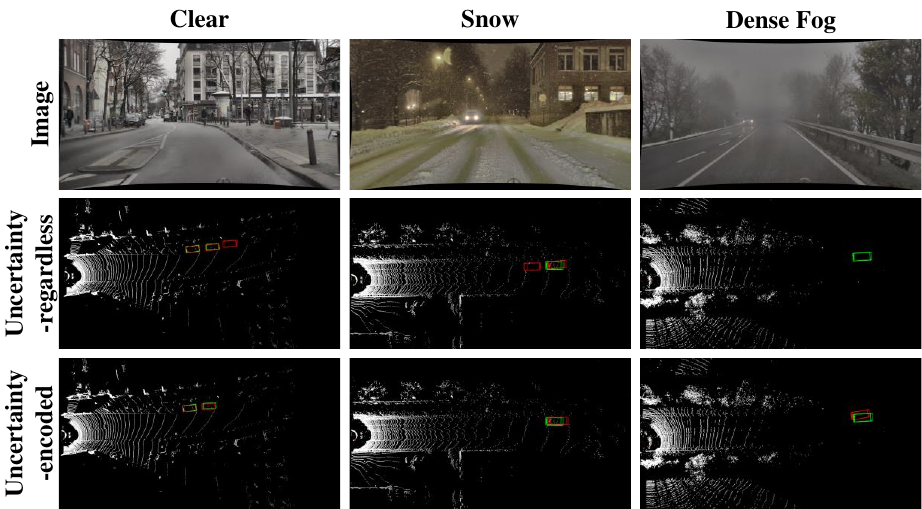}
% \vspace{0.05em}
\caption{Qualitative results of uncertainty-regardless fusion baselines, uncertainty-encoded fusion method, and the corresponding 2D images. Red bounding boxes are detections, and green bounding boxes denotes the ground truth detections.}
\label{fig:appendix_qual_res_stf}
\vspace{-1em}
\end{figure*}

\begin{figure*}[t]
\centering
\includegraphics[width=0.9\textwidth,keepaspectratio]{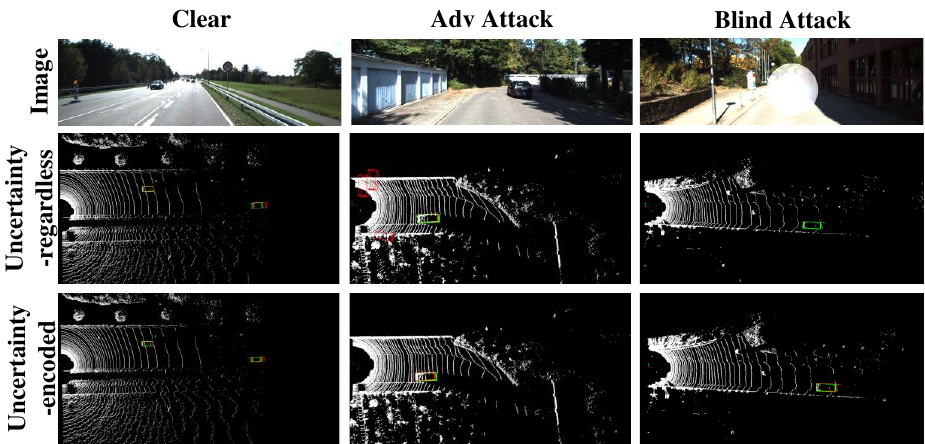}
% \vspace{0.05em}
\caption{Qualitative results of uncertainty-regardless fusion baselines, uncertainty-encoded fusion method, and the corresponding 2D images. Red bounding boxes are detections, and green bounding boxes denotes the ground truth detections.}
\label{fig:appendix_qual_res_kitti}
\vspace{-1em}
\end{figure*}

Fig.~\ref{fig:appendix_qual_res_stf} and~\ref{fig:appendix_qual_res_kitti} shows some qualitative results of uncertainty-regardless fusion baselines, uncertainty-encoded fusion method, and the corresponding 2D images across all aforementioned scenarios. Detections are represented by red bounding boxes, while ground truth detections are denoted by green bounding boxes.

%%%%%%%%%%%%%%%%%%%%%%%%%%%%%%%%%%%%%%%%%%%%%%%%%%%%%%%%%%%%%%%%%%%%%%

\end{document}